\documentclass[sigconf]{acmart}

\usepackage{amsmath}
\DeclareMathOperator*{\argmax}{arg\,max}

\usepackage{multirow}
\usepackage{makecell}
\usepackage{bbding}
\usepackage{utfsym}
\usepackage{balance}
\usepackage{subfigure}

\AtBeginDocument{%
  }

\copyrightyear{2025}
\acmYear{2025}
\setcopyright{acmlicensed}
\acmConference[MM '25]{Proceedings of the 33rd ACM International Conference on Multimedia}{October 27--31, 2025}{Dublin, Ireland}
\acmBooktitle{Proceedings of the 33rd ACM International Conference on Multimedia (MM '25), October 27--31, 2025, Dublin, Ireland}
\acmDOI{10.1145/3746027.3755297}
\acmISBN{979-8-4007-2035-2/2025/10}

\begin{document}

\title{DOMR: Establishing Cross-View Segmentation via Dense Object Matching}

\author{Jitong Liao}
\orcid{0009-0002-1097-5859}
\authornote{Both authors contributed equally to this research.}
\email{jitongliao@buaa.edu.cn}
\affiliation{%
  \institution{Hangzhou International Innovation Institute, Beihang University}
  \city{Hangzhou}
  \country{China}
}

\author{Yulu Gao}
\orcid{0000-0002-3895-1288}
\authornotemark[1]
\email{gyl97@buaa.edu.cn}
\affiliation{%
  \institution{Hangzhou International Innovation Institute, Beihang University}
  \city{Hangzhou}
  \country{China}
}

\author{Shaofei Huang}
\orcid{0000-0001-8996-9907}
\email{nowherespyfly@gmail.com}
\affiliation{%
  \institution{Faculty of Science and Technology, University of Macau}
  \city{Macau}
  \country{China}
}

\author{Jialin Gao}
\orcid{0000-0002-8554-7827}
\email{gaojialin04@meituan.com}
\affiliation{%
 \institution{Meituan}
 \city{Beijing}
 \country{China}}

\author{Jie Lei}
\orcid{0000-0003-2523-5810}
\email{jasonlei@zjut.edu.cn}
\affiliation{%
  \institution{College of Computer Science and Technology, Zhejiang University of Technology}
  \city{Hangzhou}
  \country{China}}

\author{Ronghua Liang}
\orcid{0000-0003-2077-9608}
\email{rhliang@zjut.edu.cn}
\affiliation{%
  \institution{College of Computer Science and Technology, Zhejiang University of Technology}
  \city{Hangzhou}
  \country{China}}

\author{Si Liu}
\orcid{0000-0002-9180-2935}
\email{liusi@buaa.edu.cn}
\authornote{Corresponding author.}
\affiliation{%
  \institution{School of Artificial Intelligence, Beihang University}
  \city{Beijing}
  \country{China}}

\renewcommand{\shortauthors}{Jitong Liao and Yulu Gao, et al.}

\begin{abstract}
Cross-view object correspondence involves matching objects between egocentric (first-person) and exocentric (third-person) views. It is a critical yet challenging task for visual understanding. 
In this work, we propose the Dense Object Matching and Refinement (DOMR) framework to establish dense object correspondences across views. The framework centers around the Dense Object Matcher (DOM) module, which jointly models multiple objects. Unlike methods that directly match individual object masks to image features, DOM leverages both positional and semantic relationships among objects to find correspondences.
DOM integrates a proposal generation module with a dense matching module that jointly encodes visual, spatial, and semantic cues, explicitly constructing inter-object relationships to achieve dense matching among objects. Furthermore, we combine DOM with a mask refinement head designed to improve the completeness and accuracy of the predicted masks, forming the complete DOMR framework. Extensive evaluations on the Ego-Exo4D benchmark demonstrate that our approach achieves state-of-the-art performance with a mean IoU of 49.7\% on Ego$\to$Exo and 55.2\% on Exo$\to$Ego. These results outperform those of previous methods by 5.8\% and 4.3\%, respectively, validating the effectiveness of our integrated approach for cross-view understanding.
\end{abstract}




\maketitle

\section{Introduction}

\begin{figure}[th]
    \centering
    \includegraphics[width=0.98\linewidth, trim=23.8cm 0cm 26.8cm 0cm, clip]{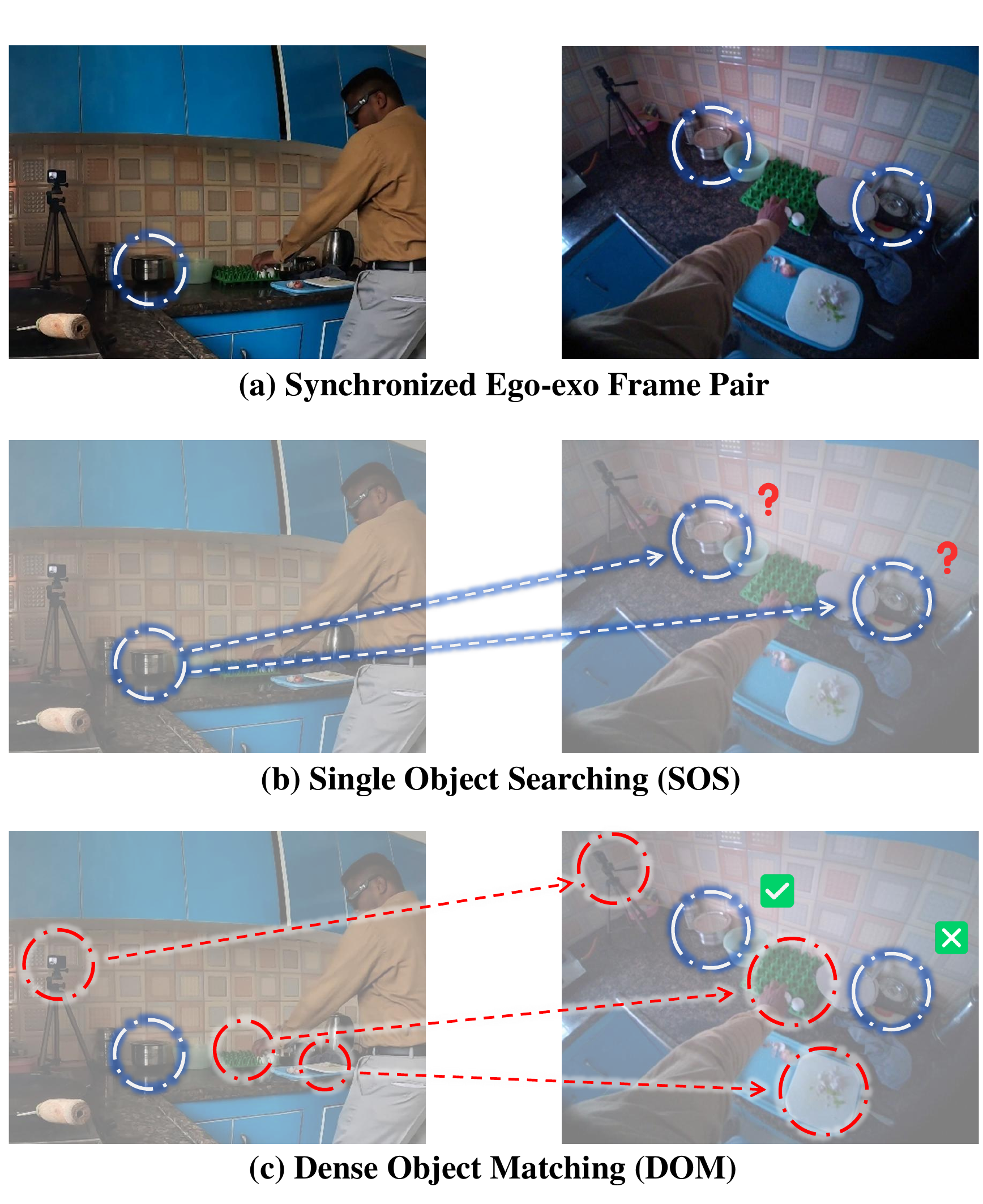}
    \caption{Comparison between Single Object Searching (SOS) and Dense Object Matching (DOM). While SOS relies on a single query and struggles with ambiguous targets, DOM performs matching among multiple objects. By leveraging spatial context from surrounding objects, DOM establishes dense correspondences suppressing incorrect matches.}
    \label{fig:sparse-vs-dense}
\end{figure}

With rapid advancements in embodied intelligence and virtual reality, aligning egocentric and exocentric views has recently emerged as a critical research direction. The core objective of the ego-exo correspondence task is to establish accurate object-level mappings between egocentric (first-person) and exocentric (third-person) viewpoints. Such an alignment enables embodied agents or users to perceive their surroundings by matching and integrating information from distinct perspectives.

Prior studies have primarily approached this challenge as a one-to-many object correspondence problem. Methods such as PSALM~\cite{zhang2024psalm} and ObjectRelator~\cite{fu2024objectrelator} leverage large language models (LLMs) to directly predict correspondences in an end-to-end manner, while others incorporate matching~\cite{kirillov2023segment} or temporal models~\cite{grauman2024ego} to capture cross-view object associations. Despite these advances, existing approaches often neglect contextual relationships among surrounding objects, leading to ambiguous matching in complex scenes. For instance, when multiple similar objects (e.g., steel bowls) appear simultaneously, existing methods struggle to establish precise correspondences without considering the spatial configuration of adjacent objects. Figure~\ref{fig:sparse-vs-dense} illustrates a scenario containing such ambiguous objects.

To overcome these limitations, we propose the \textbf{Dense Object Matching and Refinement (DOMR)} framework, a novel two-stage approach designed to exploit dense contextual information for robust ego-exo object correspondence. Instead of treating correspondence as an isolated object search problem, our \textbf{Dense Object Matcher (DOM)} module explicitly incorporates visual, spatial and semantic relationships among surrounding objects to facilitate more accurate matching. Specifically, DOM first uses an off-the-shelf open-vocabulary detector to generate bounding box proposals in both egocentric and exocentric views, then employs SAM 2~\cite{ravi2024sam} to encode these proposals into rich mask embeddings. To comprehensively represent each object, we combine positional embeddings, semantic class labels, visual features, and view-specific embeddings. These representations are further processed through transformer decoder layers to model intricate cross-view and contextual interactions, enabling effective matching even among densely populated and visually similar objects.

Moreover, considering the inherent multi-to-multi correspondence characteristics of ego-exo scenarios, we introduce a novel \textbf{Mix Matching} strategy during the post-processing stage. This strategy integrates bidirectional matching scores (ego-to-exo and exo-to-ego) to refine correspondences, significantly improving matching reliability. 
Additionally, to address the discrepancies between SAM-generated masks and manually annotated ground-truth masks, we introduce a \textbf{Mask Refinement (MR)} stage to further improve mask accuracy.
By refining selected matched proposals, our DOMR framework ensures precise and consistent segmentation results.

Extensive experiments conducted on the Ego-Exo4D benchmark~\cite{grauman2024ego} demonstrate that our DOMR framework achieves state-of-the-art performance, reaching a mean IoU of 49.7\% for Ego$\to$Exo and 55.2\% for Exo$\to$Ego tasks, outperforming previous methods by substantial margins of 5.8\% and 4.3\%, respectively. These results confirm the effectiveness of our integrated approach for cross-view visual understanding.

Our key contributions can be summarized as follows:

\begin{itemize}
    \item We introduce dense object matching, a novel correspondence strategy that leverages contextual information from surrounding objects, significantly enhancing matching accuracy compared to conventional single-object searching methods.
    \item We propose the Dense Object Matching and Refinement (DOMR) framework, comprising two complementary stages: dense object matching with the mix matching strategy and a subsequent mask refinement stage to achieve precise segmentation masks.
    \item Extensive evaluations on the Ego-Exo4D benchmark demonstrate the superior performance of DOMR, setting new state-of-the-art results on both Ego$\to$Exo and Exo$\to$Ego correspondence tasks.
\end{itemize}

\section{Related Work}

\subsection{Visual Object Correspondence}

Visual object correspondence focuses on establishing positional relationships between targets in images or videos. It is widely used in optical flow estimation, 3D reconstruction, and multi-view matching. Based on correspondence density, it can be divided into sparse matching and dense matching.

Sparse matching extracts keypoints and establishes point-to-point relationships using feature description and matching. Classical algorithms like SIFT~\cite{lowe2004distinctive} and SURF~\cite{calonder2010brief} used hand-crafted features for precise geometric descriptions, while learning-based descriptors like LIFT~\cite{yi2016lift} improved robustness. Matching typically relies on robust random sampling strategies like RANSAC~\cite{fischler1981random} to filter feature pairs and handle outliers. Recently, graph neural networks~\cite{sarlin2020superglue} with attention mechanisms have significantly enhanced performance, especially in large-baseline and non-rigid scenarios.

Dense matching focuses on per-pixel correspondence, mainly for optical flow estimation and video matching. Early methods like the Lucas-Kanade algorithm~\cite{lucas1981iterative} estimated dense fields using local smoothness assumptions, while modern approaches leverage deep learning~\cite{ummenhofer2017demon, truong2020glu} for global correlations, showing superior robustness in handling large disparities and appearance variations.

\subsection{Cross-View Modeling}

In the broad field of multi-view research, methods primarily focus on view translation~\cite{ardeshir2018egocentric, regmi2019cross, luo2025put, cheng20254diff}, novel view synthesis~\cite{liu2021infinite, ren2022look, tseng2023consistent, chan2023generative}, and aerial-ground view matching~\cite{regmi2019bridging, lin2015learning}. These works demonstrate that multi-view modeling not only involves perspective transformation but also requires consideration of scene geometry and semantic information, particularly in dynamically changing scenarios with large viewpoint variations.

Research on egocentric-exocentric modeling is still limited, largely due to the lack of high-quality, large-scale, synchronized real-world data~\cite{grauman2024ego}. However, progress has been made in related areas. Some studies focus on cross-view person matching~\cite{ardeshir2016ego2top, xu2018joint, fan2017identifying, wen2021seeing}, while others explore learning view-invariant~\cite{ardeshir2018egocentric, sermanet2018time, yu2019see, xue2023learning} or egocentric-specific features~\cite{li2021ego}. Additionally, some research targets perspective translation. Exo2Ego~\cite{luo2025put} transforms third-person videos into first-person perspectives using high-level structure transformation and diffusion-based pixel-level detail completion, emphasizing hand manipulation. While these works provide theoretical foundations for understanding cross-view relationships, practical applications remain constrained by limited data scale and diversity.

\subsection{Segmentation Models}

Traditional segmentation research has primarily focused on semantic segmentation~\cite{chen2014semantic,chen2017rethinking,chen2017deeplab,chen2018encoder}, instance segmentation~\cite{hafiz2020survey,liu2018path,bolya2019yolact}, and panoptic segmentation~\cite{kirillov2019panoptic}. However, these models are typically task-specific, leading to limited generalization~\cite{wang2023large, zhang2023comprehensive}. Several efforts have explored unified frameworks like K-Net~\cite{zhang2021k}, MaskFormer~\cite{cheng2021per}, and Mask2Former~\cite{cheng2022masked}.

Recent work emphasizes task-agnostic fundamental segmentation models pretrained on large-scale datasets for strong adaptability. The Segment Anything Model (SAM)~\cite{kirillov2023segment} stands out with comprehensive image segmentation capabilities and remarkable zero-shot performance. SAM supports interactive segmentation via prompts like boxes and points and can segment all objects in an image. SAM 2~\cite{ravi2024sam} extends this to video segmentation while improving speed. SegGPT~\cite{wang2023seggpt} unifies tasks through in-context learning, and SEEM~\cite{zou2023segment} incorporates multimodal prompts, including visual, textual, and reference cues. PSALM~\cite{zhang2024psalm} uses Large Multimodal Models (LMMs) for diverse tasks within a unified framework. Recent studies also explore unified detection models like Grounding DINO~\cite{liu2024grounding}, YOLO-World~\cite{Cheng2024YOLOWorld}, YOLO-UniOW~\cite{liu2024yolouniow}, and YOLO-E~\cite{wang2025yoloerealtimeseeing}, which can prompt segmentation models, highlighting the potential of unified tasks.

\begin{figure*}[h]
    \centering
    \includegraphics[width=\linewidth, trim=2cm 2cm 1cm 1.8cm, clip]{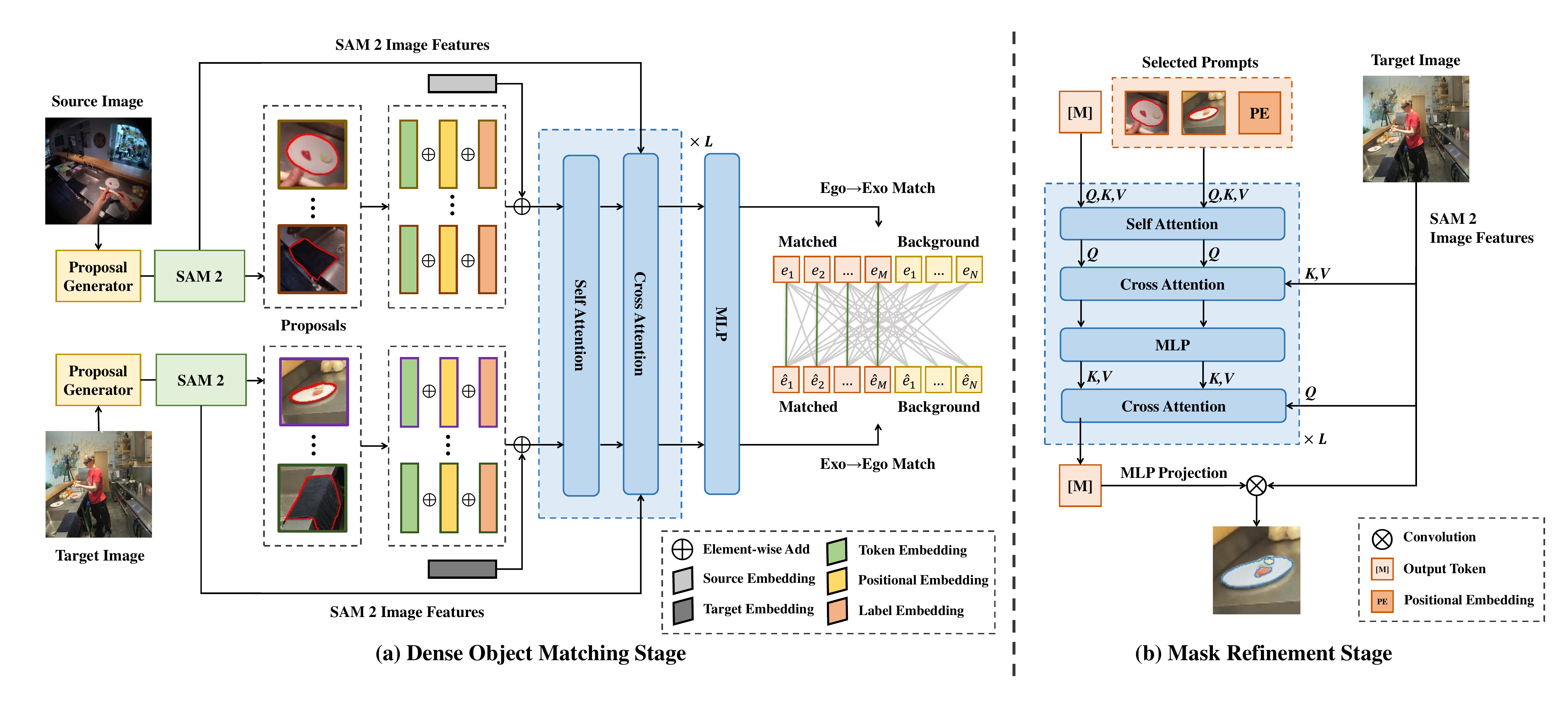}
    \caption{Overview of the DOMR architecture, where we take Ego$\to$Exo as an example. DOMR is trained in two stages: (a) The Dense Object Matching (DOM) stage processes mask proposals from the proposal generator and SAM, where embeddings interact with each other and SAM's image features. Contrastive supervision is only applied to the best ground-truth-matched proposal. (b) The Mask Refinement (MR) stage employs learnable output tokens prompted with cross-view embeddings and SAM's positional encodings, which undergo two-way interaction with image features. Final masks are produced via MLP processing and convolutional operations on the image features.}
    \label{fig:pipeline}
\end{figure*}

\section{Method}

\subsection{Task Definition}

The ego-exo correspondence task aims to establish object-level relationships between egocentric and exocentric views. Given paired source and target videos $V^{\text{src}}, V^{\text{tgt}}$ derived from synchronized ego-exo videos, along with a set of masks $M^{\text{src}}$ representing objects of interest in each frame of the source video, the objective is to identify corresponding masks $M^{\text{tgt}}$ for the same objects in $V^{\text{tgt}}$. The source and target videos can be either ego- or exo-views: we denote the task as \textbf{Ego$\to$Exo} when the source is an egocentric video and target is exocentric, and \textbf{Exo$\to$Ego} for the reverse configuration.

This task originates from the ego-exo correspondence benchmark proposed in the Ego-Exo4D dataset~\cite{grauman2024ego}. The model is expected to focus on objects of interest across viewpoints without additional inputs such as semantic labels, object names, camera intrinsics/extrinsics, IMU or sensor data, which is typically unavailable in consumer-grade camera devices.

\begin{table}[tb]
    \caption{Coverage of proposal generator.}
    \label{tab:coverage-of-proposal-generator}
    \centering
    \begin{tabular}{cccc}
        \toprule
        \multicolumn{2}{c}{\textbf{Ego-View}} & \multicolumn{2}{c}{\textbf{Exo-View}} \\
        \textbf{Box IoU} & \textbf{Mask IoU} & \textbf{Box IoU} & \textbf{Mask IoU} \\
        \cmidrule(lr){1-2} \cmidrule(lr){3-4}
        78.5 & 77.4 & 71.5 & 67.1 \\
        \bottomrule
    \end{tabular}
\end{table}

\subsection{Overview}

Unlike traditional one-to-one segmentation methods, our proposed DOMR focuses on establishing dense correspondences of objects across different viewpoints. This approach leverages objects' positional relationships and contextual associations across views, rather than relying solely on image feature matching. As shown in Figure~\ref{fig:pipeline}, DOMR consists of two stages: the dense object matching stage and the mask refinement stage.

For synchronized ego-exo video pairs, we process each source-target image frame sequentially. In the dense object matching stage, a proposal generator first extracts multiple bounding boxes from both views as candidate regions, which are then fed into SAM 2 to produce corresponding candidate masks. 
The representation of each proposal integrates three complementary embeddings: a token embedding encoding visual features, a positional embedding capturing spatial relationships, and a label embedding incorporating semantic information, which are subsequently fused and transformed through an attention module to produce object embeddings.
We then compare these embeddings across the two views to find the most similar pairs, which we treat as corresponding objects. We refer to the modules in this stage as the Dense Object Matcher (DOM). In the mask refinement stage, an additional Mask Refinement (MR) head integrates information from both views to correct errors from the first stage, yielding more complete and accurate target masks.

\subsection{Proposal Generation}

To obtain object mask proposals from each frame, one option is to use segmentation models like SAM~\cite{kirillov2023segment} or SAM 2~\cite{ravi2024sam} to automatically generate masks.
However, SAM 2's automatic mask generator efficiency depends on sampling density (e.g., $32\times32$ grid points), where increased sampling improves proposal quality at the cost of computational overhead, making accuracy-efficiency trade-offs challenging.

For bounding box-level proposals, open vocabulary object detectors (e.g. YOLO-World~\cite{Cheng2024YOLOWorld}, YOLO-UniOW~\cite{liu2024yolouniow}, YOLO-E~\cite{wang2025yoloerealtimeseeing}) offer efficient multi-object localization, which can be fed into SAM to derive corresponding masks. 
Notably, the number of salient objects per image typically remains below SAM's default sampling density, making box prompts significantly more computationally efficient than dense point prompts.

Based on these considerations, DOM employs a pretrained YOLO-UniOW detector as its proposal generator. Given an input frame $\mathcal{I}$, YOLO-UniOW generates $N$ bounding box proposals using the LVIS vocabulary~\cite{gupta2019lvis}. These proposals then serve as box prompts for SAM 2 to predict corresponding object masks, ultimately producing a set of high-quality mask-level proposals $\{m_1, m_2, \ldots, m_N\}$.

However, inaccuracies in the proposal generation step can lead to performance degradation, since the model may not generate a precise mask proposal for every ground-truth object. To measure the prediction bias, we pair each ground truth with the mask that obtains the greatest IoU with it in the Ego-Exo4D training set, indicating proposal coverage. Table~\ref{tab:coverage-of-proposal-generator} shows the proposal coverage in terms of IoU between generated proposals and ground-truth masks.
With average IoUs above 70\% in both ego and exo views, this proposal generator successfully localizes most objects via its bounding box outputs.
For mask prediction, the coverage remains above 65\%, indicating that most bounding boxes can be covered with suitable masks. The maximum coverage also indicates the upper bound precision of the proposed DOM. The goal of DOM is to establish dense relationships across objects, whether the object is considered ground truth or not; therefore, the coverage remains acceptable for our dense matching approach.

\subsection{Dense Object Matching}
\label{sec:dense-object-matching}

Our proposed DOM module specializes in aligning mask proposals from paired video frames while learning dense matching without camera parameters or sensor data. We observe that positional information provides meaningful guidance for cross-view object matching. When multiple objects appear close in one view, they should maintain similar relative positions in another view. This spatial consistency forms the basis of our dense matching approach. For each object, we utilize the bounding box coordinates $b = [x_1, y_1, x_2, y_2]$ predicted by the proposal generator to construct positional embeddings. The corner points $(x_1, y_1)$ and $(x_2, y_2)$ are encoded into positional embeddings $\mathrm{PE}(x_1, y_1)$ and $\mathrm{PE}(x_2, y_2)$ respectively. Their combined representation $e^{\text{pos}} = \mathrm{PE}(x_1, y_1) + \mathrm{PE}(x_2, y_2)$ captures the object's spatial information.

While dense matching primarily relies on positional relationships, position-independent sparse features also contribute to dense cross-view object association:
1) \textbf{Token Embedding}. In SAM, each prompt generates token embeddings $e^{\text{token}}$ through attention mechanisms, which are then transformed via MLP into convolutional kernels that operate on upsampled image features, ultimately predicting output masks. Kernels activate related regions on the feature map, and consequently embrace potential mask-level visual features crucial for object matching.
2) \textbf{Label Embedding}. YOLO-UniOW employs the CLIP~\cite{radford2021learning} text encoder to assign class labels to objects. The matched label embeddings $e^{\text{label}}$ encode category-related semantic information, enabling cross-view object category comparison.

Finally, all embeddings are aggregated through a linear projection to form comprehensive object representations combining both positional and non-positional information in $d$ dimensions:
\begin{equation}
    e = \text{Linear}(e^{\text{token}}) + \text{Linear}(e^{\text{pos}}) + \text{Linear}(e^{\text{label}}).
\end{equation}
The DOM module employs $L$-layer attention blocks to model interactions between objects. Given $N$ proposals generated from both frames $\mathcal{I}^{\text{src}}$ and $\mathcal{I}^{\text{tgt}}$ respectively, the $l$-th attention block processes the concatenated object embeddings $E^{(0)} \in \mathbb{R}^{2N \times d}$, through:
\begin{equation}
    E^{(l)}_{\text{sa}} = \text{SA}^{(l)}(E^{(l-1)} + \text{PE}_{\text{emb}}, E^{(l-1)} + \text{PE}_{\text{emb}}, E^{(l-1)}) + E^{(l-1)},
\end{equation}
where $\text{SA}^{(l)}$ denotes the $l$-th multi-head self-attention module, and $\text{PE}_{\text{emb}}$ is a learnable parameter indicating the view that each embedding comes from. The results are further processed via cross-attention with concatenated SAM 2 image features $\mathcal{F} \in \mathbb{R}^{2HW \times D}$:
\begin{equation}
    E^{(l)}_{\text{ca}} = \text{CA}(E^{(l)}_{\text{sa}} + \text{PE}_{\text{emb}}, \mathcal{F} + \text{PE}_{\text{img}}, \mathcal{F}) + E^{(l)}_{\text{sa}},
\end{equation}
\begin{equation}
    E^{(l)} = \text{FFN}(E^{(l)}_{\text{ca}}) + E^{(l)}_{\text{ca}},
\end{equation}
where $\text{CA}^{(l)}$ denotes the $l$-th multi-head cross-attention module, and $\text{PE}_{\text{img}}$ represents image positional embedding.
The final ego- and exo-view embeddings are transformed through linear layers to produce object embeddings $e \in \mathbb{R}^{d}$ incorporating both dense and sparse information. We use $e$ and $\hat{e}$ to represent the embeddings of the source and target views, respectively. For clarity in differentiating the variables associated with the source and the target view in this paper, target variables are denoted with a hat mark, whereas source variables are left unmarked. For any two object embeddings $e_i$ and $\hat{e}_j$, their similarity is computed as:
\begin{equation}
    \label{eq:embed-sim}
    \mathrm{sim}(e_i, \hat{e}_j) = \frac{e_i \cdot \hat{e}_j}{\|e_i\|\|\hat{e}_j\|}.
\end{equation}

\subsection{Matching Strategy}

DOM stage produces a similarity matrix between proposals from ego-view and exo-view, requiring appropriate matching strategy $\sigma$ applied on it. For arbitrary object in the source view, it is natural to select the best-matched proposal with the highest similarity in the target view:
\begin{equation}
    \sigma^{\text{top1}}(i) = \argmax_{j=1}^N \ \mathrm{sim}(e_i, \hat{e}_j),
\end{equation}
and denote it ``top-1 strategy''. However, establishing correspondence from ego-view to exo-view may be more challenging than the reverse direction, as the exo-view typically contains more objects and the objects may be significantly smaller than those in the ego-view. We address this issue by introducing the property that object should keep symmetric consistency~\cite{feng2024unveiling} across views: for two objects X and Y in different views, if Y is the best match for X, then X should likewise be the best match for Y.
Specifically, we apply the softmax function to the similarity matrix along both horizontal and vertical directions, representing the matching probability of each directed connection, respectively:

\begin{equation}
    \mathrm{sim}^{\text{src}}(e_i, \hat{e}_j) = \frac {\exp [\tau \cdot \mathrm{sim}(e_i, \hat{e}_j)]} {\sum_{j=1}^{N} \exp [\tau \cdot \mathrm{sim}(e_i, \hat{e}_j) ]},  
\end{equation}    
\begin{equation}
    \mathrm{sim}^{\text{tgt}}(e_i, \hat{e}_j) = \frac {\exp [\tau \cdot \mathrm{sim}(e_i, \hat{e}_j)]} {\sum_{i=1}^{N} \exp [\tau \cdot \mathrm{sim}(e_i, \hat{e}_j) ]},  
\end{equation}
where $\tau$ is a learnable temperature parameter during training in Equation~\ref{eq:celoss-src} and Equation~\ref{eq:celoss-tgt}. The symmetric consistency score for each object pair is computed as the product of their Ego$\to$Exo and Exo$\to$Ego score in the similarity matrix, with the pair achieving the maximum product selected as the best match:
\begin{equation}
    \sigma^{\text{mix}}(i) = \argmax_{j=1}^N \ \mathrm{sim}^{\text{src}}(e_i, \hat{e}_j) \cdot \mathrm{sim}^{\text{tgt}}(e_i, \hat{e}_j).
\end{equation}
We refer to this strategy as the ``mix strategy'', which is set as default strategy in DOM. 

\subsection{Mask Refinement}

While we can directly select the proposal mask in various strategies as the correspondence result, inevitable discrepancies exist between proposals and ground truth masks. 
This occurs because proposal hierarchies may introduce misalignment such as selecting only part of the object, and the predicted masks may contain boundary biases compared to manual annotations.
Table~\ref{tab:coverage-of-proposal-generator} shows the average IoU between ground truth masks and proposal masks. To correct bias introduced by the proposal generator, we further propose the Dense Object Matching and Refinement (DOMR) to refine matched masks and provide enhanced results based on the image.

The DOMR combines DOM with an extra Mask Refinement (MR) head, which contains multiple layers of two-way transformer blocks that interact bidirectionally from query prompts to image and vice versa, consistent with SAM 2 mask decoder. For each query, MR utilizes a learnable output token for predicting features related to refined masks, followed by source object embedding $e$ and the best matched object embedding $\hat{e}$ from DOM. Additionally, the query is attached with target view positional embeddings $\mathrm{PE}(x_1, y_1)$ and $\mathrm{PE}(x_2, y_2)$ introduced in Section~\ref{sec:dense-object-matching} to specify positional information of the mask. 
The two-way transformer block processes queries through four sequential stages: self-attention on queries, query-to-image cross-attention, MLP transformation, and image-to-query cross-attention, with residual connections and layer normalization applied after each operation. The output token is passed through an MLP layer to serve as convolutional kernels and perform $1\times1$ convolution with upsampled image features to predict low-resolution masks, which eventually produce the refined segmentation.

\subsection{Loss Functions}

\textbf{Training DOM}. The loss function only contains a contrastive loss for object embedding to facilitate cross-view feature fusion. For all $M$ objects in frame $\mathcal{I}$, let $\sigma(i), \hat{\sigma}(i)$ denote the index of the proposal mask with maximum IoU to the $i$-th ground truth mask $m_i^{\text{gt}}$ and $\hat{m}_i^{\text{gt}}$ in both views:
\begin{equation}
\sigma(i) = \argmax_{j=1}^{N} \ \mathrm{IoU}(m_i^{\text{gt}}, m_j),
\end{equation}
\begin{equation}
\hat{\sigma}(i) = \argmax_{j=1}^{N} \ \mathrm{IoU}(\hat{m}_i^{\text{gt}}, \hat{m}_j).
\end{equation}
This indicates that the $\sigma(i)$-th source object and $\hat{\sigma}(i)$-th target object correspond to the $i$-th ground truth pair. We optimize model by maximizing their similarity while minimizing similarities to other targets using cross-entropy loss:
\begin{equation}
\label{eq:celoss-src}
\mathrm{CE}^{\text{src}}(i) = - \log \frac {\exp [\tau \cdot \mathrm{sim}(e_{\sigma(i)}, \hat{e}_{\hat{\sigma}(i)}) ]} {\sum_{j=1}^{N} \exp [\tau \cdot \mathrm{sim}(e_{\sigma(i)}, \hat{e}_j) ]},
\end{equation}
\begin{equation}
\label{eq:celoss-tgt}
\mathrm{CE}^{\text{tgt}}(i) = - \log \frac {\exp [\tau \cdot \mathrm{sim}(e_{\sigma(i)}, \hat{e}_{\hat{\sigma}(i)}) ]} {\sum_{j=1}^{N} \exp [\tau \cdot \mathrm{sim}(e_j, \hat{e}_{\hat{\sigma}(i)}) ]}.
\end{equation}
Due to detector limitations, proposals might not cover all objects, and low-IoU samples could harm training. Thus, we only select proposals with IoU > 0.5 in both views for contrastive learning and average the results:
\begin{equation}
\mathcal{L}_{\text{DOM}} = \frac 12 \sum_{i=1}^{M} \left[ \mathrm{CE}^{\text{src}}(i) + \mathrm{CE}^{\text{tgt}}(i) \right] \cdot v(i),
\end{equation}
where 
\begin{equation}
    v(i) =
    \begin{cases}
        1   & \text{if } \mathrm{IoU}(m_i^{\text{gt}}, m_{\sigma(i)}) \text{ and } \mathrm{IoU}(\hat{m}_i^{\text{gt}}, \hat{m}_{\hat{\sigma}(i)}) > 0.5 \\
        0   & \text{otherwise} \\
    \end{cases}
    .
\end{equation}

During inference, the ground truth masks do not directly participate in DOM to keep proposal consistency with training. Instead, their bounding box is served as box prompts for SAM 2 to produce corresponding proposals, concatenated with original proposals for matching. 

\noindent\textbf{Training DOMR}. DOM is frozen during this stage. The loss function incorporates mask prediction losses from the ego-exo mask prediction head, including focal loss~\cite{lin2017focal} and dice loss~\cite{milletari2016v}. The loss is computed with $ M_{\text{pos}} = \sum_{i=1}^{M} v(i)$:
\begin{equation}
\mathcal{L}_{\text{focal}} = \frac{1}{ M_{\text{pos}} } \sum_{i=1}^{M} v(i) \cdot \text{FocalLoss}(m_i^{\text{pred}}, m_i^{\text{gt}}),
\end{equation}

\begin{equation}
\mathcal{L}_{\text{dice}} = \frac{1}{ M_{\text{pos}} } \sum_{i=1}^{M} v(i) \cdot \text{DiceLoss}(m_i^{\text{pred}}, m_i^{\text{gt}}).
\end{equation}
The final loss is linear combination of $\mathcal{L}_{\mathrm{focal}}$ and $\mathcal{L}_{\mathrm{dice}}$:
\begin{equation}
    \mathcal{L}_{\mathrm{DOMR}} = \lambda_{\text{focal}} \cdot \mathcal{L}_{\mathrm{focal}} + \lambda_{\text{dice}} \cdot \mathcal{L}_{\mathrm{dice}}.
\end{equation}

\begin{table*}[!t]
    \caption{Comparison with previous state-of-the-art methods on ego-exo correspondence benchmark. ZSL denotes zero-shot learning result. Type ``S'' means spatial-only modeling, while ``ST'' means spatio-temporal modeling.}
    \label{tab:main-result}
    \begin{tabular}{lccc|lccc}
        \toprule
        \multicolumn{4}{c|}{\textbf{Ego$\to$Exo}} & \multicolumn{4}{c}{\textbf{Exo$\to$Ego}} \\
        \textbf{Method} & \textbf{ZSL} & \textbf{Type} & \textbf{IoU} $\uparrow$ & \textbf{Method} & \textbf{ZSL} & \textbf{Type} & \textbf{IoU} $\uparrow$ \\
        \midrule
        XSegTx~\cite{grauman2024ego} & $\checkmark$ & S & 0.3 & XSegTx~\cite{grauman2024ego} & $\checkmark$ & S & 1.3 \\
        XSegTx~\cite{grauman2024ego} & $\times$ & S & 6.2 & XSegTx~\cite{grauman2024ego} & $\times$ & S & 30.2 \\
        XView-Xmem~\cite{grauman2024ego} & $\checkmark$ & ST & 16.2 & XView-Xmem~\cite{grauman2024ego} & $\checkmark$ & ST & 13.5 \\
        XView-Xmem~\cite{grauman2024ego} & $\times$ & ST & 17.7 & XView-Xmem~\cite{grauman2024ego} & $\times$ & ST & 20.7 \\
        XView-Xmem + XSegTx~\cite{grauman2024ego} & $\times$ & ST & 36.9 & XView-Xmem + XSegTx~\cite{grauman2024ego} & $\times$ & ST & 36.1 \\
        SEEM~\cite{zou2023segment} & $\checkmark$ & S & 1.1 & SEEM~\cite{zou2023segment} & $\checkmark$ & S & 4.1 \\
        PSALM~\cite{zhang2024psalm} & $\checkmark$ & S & 7.9 & PSALM~\cite{zhang2024psalm} & $\checkmark$ & S & 9.6 \\
        PSALM~\cite{zhang2024psalm} & $\times$ & S & 41.3 & PSALM~\cite{zhang2024psalm} & $\times$ & S & 47.3 \\
        ObjectRelator~\cite{fu2024objectrelator} & $\times$ & S & 43.9 & ObjectRelator~\cite{fu2024objectrelator} & $\times$ & S & 50.9 \\
        \midrule
        \textbf{DOMR (Ours)} & $\times$ & S & \textbf{49.7} & \textbf{DOMR (Ours)} & $\times$ & S & \textbf{55.2} \\
        \bottomrule
    \end{tabular}
\end{table*}

\section{Experiments}

\subsection{Preparation}

\textbf{Dataset}. We conduct evaluation on ego-exo correspondence benchmark from Ego-Exo4D dataset~\cite{grauman2024ego}, a large-scale collection of synchronized egocentric and third-person videos capturing professional skill demonstrations across diverse domains. It embraces manually annotated 1,335 takes, 5,566 target objects with 742K egocentric masks and 1.1M exocentric masks. All 1.8 million masks are sampled from takes at 1 FPS, and each take possesses an average annotations for 5.5 objects tracked across 173 frames. The labeled objects mainly consist of wearer-interacted objects, activity-relevant environmental objects, and body parts. 
We use the standard train/val split provided by Ego-Exo4D. The evaluation metric is the mean IoU between the predicted masks and ground-truth masks.

\noindent\textbf{Implementation Details}. 
No data augmentation is applied during training. For our DOM  stage, we use pretrained YOLO-UniOW-L~\cite{wang2024yolov10, liu2024yolouniow} for proposal generator and use SAM 2.1 Hiera-B+~\cite{ravi2024sam} for feature extraction. The number of attention blocks $L$ is set to 6. We adopt the AdamW~\cite{loshchilov2017decoupled} optimizer with an initial learning rate of $1\times 10^{-4}$, and weight decay is set to $1 \times 10^{-4}$. The model is trained for $800,000$ iterations, while the learning rate decayed by a factor of 0.1 after $533,336$ and $733,337$ iteration, respectively. For our DOMR model, we initialize the model with the mask decoder from SAM 2.1 Hiera-B+. The initial learning rate is reduced to $5\times 10^{-5}$, and the other settings is kept as that for DOM. Following SAM~\cite{kirillov2023segment}, we set $\lambda_{\text{focal}} = 20$ and $\lambda_{\text{dice}} = 1$. All experiments are conducted on $4\times$ NVIDIA A800 GPUs with a total batch size of $16$.

\subsection{Main Results}
We evaluate our proposed DOMR alongside existing competitive methods on the Ego-Exo4D validation split, following the established evaluation protocols. As illustrated in Table~\ref{tab:main-result}, our DOMR achieves new state-of-the-art performance, obtaining IoU scores of 49.7\% for the Ego$\to$Exo task and an impressive 55.2\% for the Exo$\to$Ego task.
For reference, XView-XMem + XSegTx, a recently proposed method, achieves 36.9\% (Ego$\to$Exo) and 36.1\% (Exo$\to$Ego).
It is noteworthy that XView-Xmem + XSegTx employs spatio-temporal (ST) modeling, leveraging both spatial and temporal information, theoretically benefiting from richer context. In contrast, our DOMR employs a simpler spatial-only (S) configuration, avoiding temporal modeling altogether. 
Even without temporal cues, DOMR outperforms XView-Xmem + XSegTx by a large margin—about +12.8 and +19.1 points on Ego$\to$Exo and Exo$\to$Ego, respectively.
Moreover, compared with the previously leading state-of-the-art method, ObjectRelator, which achieved IoUs of 43.9\% (Ego$\to$Exo) and 50.9\% (Exo$\to$Ego), our DOMR demonstrates significant improvements of 5.8\% and 4.3\%, respectively. These advancements underscore the effectiveness of DOMR's novel dense object matching and mask refinement modules.

\subsection{Ablation Studies}

\noindent\textbf{Component Analysis.} 
As shown in  Table~\ref{tab:abl-component-analysis}, We conduct comprehensive ablation experiments to validate the effectiveness of each module in our proposed framework. First, by utilizing our Dense Object Matching stage (DOM) without the mix matching strategy, we achieve IoU scores of 47.2\% (Ego$\to$Exo) and 53.3\% (Exo$\to$Ego), surpassing the previous state-of-the-art method ObjectRelator, which reported 43.9\% (Ego$\to$Exo) and 50.9\% (Exo$\to$Ego). Incorporating our proposed mix strategy into DOM further improves the performance, resulting in IoU increases to 49.0\% and 53.4\% for Ego$\to$Exo and Exo→Ego respectively, clearly demonstrating the effectiveness of mix strategy. Finally, the Mask Refinement (MR) stage provides an additional significant enhancement, further raising the IoU scores to new state-of-the-art results, validating that each proposed module significantly contributes to our overall model performance.

\noindent\textbf{Impact of Proposal Quantity.} 
Table~\ref{tab:abl-num-proposals} illustrates how the number of proposals generated affects overall performance. We conducted experiments using 80, 120, and 160 proposals as inputs to the DOM stage, and results clearly demonstrate that increasing the number of proposals enhances performance. The reason for this improvement is that the Ego-Exo4D dataset contains many small target objects, and since the general proposal generator is not explicitly optimized for detecting specific categories, producing more proposals effectively improves coverage. Another possible contributing factor is that more proposals may introduce richer dense matching information, potentially aiding in establishing more complete correspondences.

\noindent\textbf{Impact of Query Components}. 
We investigate how different matching components affect embedding generation performance in the DOM attention model. Results are shown in Table~\ref{tab:abl-stage1-multimodal}. 
The ``pos'' component refers to bounding box corner points position embeddings, while the ``label'' represents label embeddings obtained by processing LVIS vocabulary detection results through a CLIP text encoder (from proposal generator). 
The results indicate that compared to using both token and positional embeddings, incorporating text embeddings improves IoU by 1.7\% on Ego$\to$Exo and 2.1\% on Exo$\to$Ego, demonstrating label embedding's effectiveness in distinguishing identical or different objects across views. 
Following SAM~\cite{kirillov2023segment}, we also explored the effect of latent space in SAM, denoted as ``feat'' in the table. We extract multi-level features from the first two backbone layers in SAM 2, cropping and averaging these features to create composite image embeddings that capture visual characteristics at different scales as feature embeddings. However, the participation of image feature embeddings does not improve performance and instead introduces additional learning difficulties on Exo$\to$Ego task. This is because the token embeddings already interact with the image and contain visual features of corresponding objects, making the redundant feature embeddings unable to provide further performance gains.

\begin{table}[tb]
    \caption{Component analysis. MIX denotes mix strategy.}
    \label{tab:abl-component-analysis}
    \centering
    \begin{tabular}{lcc}
        \toprule
        \multirow{2}{*}{\raisebox{-0.6ex}{\textbf{Method}}}  & \multicolumn{2}{c}{\textbf{IoU $\uparrow$}} \\
        \cmidrule(lr){2-3}
          & \textbf{Ego$\to$Exo} & \textbf{Exo$\to$Ego} \\
        \midrule
        DOM (w/o MIX)  & 47.2  & 53.3\\
        DOM  & 49.0 & 53.4 \\
        DOM $+$ MR & \textbf{49.7} & \textbf{55.2} \\
        \bottomrule
    \end{tabular}
\end{table}

\begin{table}[tb]
    \caption{Impact of proposal quantity.}
    \label{tab:abl-num-proposals}
    \centering
    \begin{tabular}{cccc}
        \toprule
        \multirow{2}{*}{\raisebox{-0.6ex}{\textbf{Method}}} & \multirow{2}{*}{\raisebox{-0.6ex}{\textbf{\# of Proposals}}} & \multicolumn{2}{c}{\textbf{IoU $\uparrow$}} \\
        \cmidrule(lr){3-4}
         & & \textbf{Ego$\to$Exo} & \textbf{Exo$\to$Ego} \\
        \midrule
        DOM & 80 & 46.0 & 52.6 \\
        DOM & 120 & 48.6 & 52.6 \\
        DOM & 160 & \textbf{49.0} & \textbf{53.4} \\
        \bottomrule
    \end{tabular}
\end{table}

\begin{table}[tb]
    \caption{Impact of query components.}
    \label{tab:abl-stage1-multimodal}
    \centering
    \begin{tabular}{cccccc}
        \toprule
        \multirow{2}{*}{\raisebox{-0.6ex}{\textbf{Method}}} & \multicolumn{3}{c}{\textbf{Component}} & \multicolumn{2}{c}{\textbf{IoU $\uparrow$}} \\
        \cmidrule(lr){2-4} \cmidrule(lr){5-6}
         & \textbf{pos} & \textbf{label} & \textbf{feat} & \textbf{Ego$\to$Exo} & \textbf{Exo$\to$Ego} \\
        \midrule
        DOM & \checkmark &  &  & 47.3 & 51.3 \\
        DOM &  & \checkmark &  & 48.5 & 53.0 \\
        DOM & \checkmark & \checkmark &  & 49.0 & \textbf{53.4} \\
        DOM & \checkmark & \checkmark & \checkmark & \textbf{49.1} & 53.3 \\
        \bottomrule
    \end{tabular}
\end{table}

\begin{table}[tb]
    \caption{Impact of dense object matching design. MIX denotes mix strategy.}
    \label{tab:abl-stage1-match}
    \centering
    \begin{tabular}{lccc}
        \toprule
        \multirow{2}{*}{\raisebox{-0.6ex}{\textbf{Method}}} & \multirow{2}{*}{\raisebox{-2ex}{\textbf{\makecell{Capable\\with MIX}}}} & \multicolumn{2}{c}{\textbf{IoU $\uparrow$}} \\
        \cmidrule(lr){3-4}
         &  & \textbf{Ego$\to$Exo} & \textbf{Exo$\to$Ego} \\
        \midrule
        SOS & $\times$ & 45.1 & 46.0 \\
        DOM (w/o MIX) & \checkmark & 47.2 & 53.3 \\
        DOM & \checkmark & \textbf{49.0} & \textbf{53.4} \\
        \bottomrule
    \end{tabular}
\end{table}

\noindent\textbf{Impact of Dense Object Matching Design}. 
To investigate the effectiveness of dense object matching design, we also train the model using single object searching. Compared to the original dense matching approach, SOS uses only the best-matched proposal as source query while keeping the target queries consistent with DOM. As shown in Table~\ref{tab:abl-stage1-match}, employing only a single proposal as query degrades an 2.1\% mIoU performance in Ego$\to$Exo settings and 7.3\% in Exo$\to$Ego, as target proposals cannot obtain dense spatial clues about objects from a single source and can only rely on feature for matching, leading to worse performance. 
It is worth noting that our proposed mix matching strategy specifically relies on this dense object matching design and is not applicable to single object matching design.

\noindent\textbf{Impact of Matching Strategy}. 
Table~\ref{tab:abl-sampling-strategy} presents the performance of DOM with two different matching strategies. The results demonstrate that the mix strategy outperforms directly selecting the top-1 match with the highest similarity, further improving IoU by 1.8 on Ego$\to$Exo, benefiting from the matching results from exo-view to ego-view, and 0.1 on Exo$\to$Ego. This method fully utilizes the bidirectional relationship characteristics of dense matching, which cannot be employed in single object matching approach. Additional experiment with the Hungarian matching algorithm, a dense matching approach as well, reveals its limitations in this scenario, as not every object can be matched in another view, eventually disturbing the matching result. 

\noindent\textbf{Impact of Joint Training}. 
We conduct experiments to investigate the joint training effect of DOM within the DOMR framework. Table~\ref{tab:abl-freeze-dom} compares the performance when DOM is either frozen or jointly trained with MR. The results demonstrate that joint training disturbs DOM's crucial role in predicting rough segmentations. By freezing DOM during MR training, the refinement module can focus solely on improving proposals, ultimately achieving higher mIoU.

\begin{table}[tb]
    \caption{Impact of matching strategy. TOP1 and MIX denote top-1 strategy and mix strategy, separately.}
    \label{tab:abl-sampling-strategy}
    \centering
    \begin{tabular}{cccc}
        \toprule
        \multirow{2}{*}{\raisebox{-0.6ex}{\textbf{Method}}} & \multirow{2}{*}{\raisebox{-0.6ex}{\textbf{Strategy}}} & \multicolumn{2}{c}{\textbf{IoU $\uparrow$}} \\
        \cmidrule(lr){3-4}
         &  & \textbf{Ego$\to$Exo} & \textbf{Exo$\to$Ego} \\
        \midrule
        DOM & TOP1 & 47.2 & 53.3 \\
        DOM & Hungarian & 28.7 & 35.9 \\
        DOM & MIX & \textbf{49.0} & \textbf{53.4} \\
        \bottomrule
    \end{tabular}
\end{table}

\begin{table}[tb]
    \caption{Impact of joint training of DOM. $^\dagger$ indicates DOM is frozen during training.}
    \label{tab:abl-freeze-dom}
    \centering
    \begin{tabular}{lcccc}
        \toprule
        \multirow{2}{*}{\raisebox{-0.6ex}{\textbf{Method}}} & \multicolumn{2}{c}{\textbf{DOM stage IoU $\uparrow$}} & \multicolumn{2}{c}{\textbf{MR stage IoU $\uparrow$}} \\
        \cmidrule(lr){2-3} \cmidrule(lr){4-5}
         & \textbf{Ego$\to$Exo} & \textbf{Exo$\to$Ego} & \textbf{Ego$\to$Exo} & \textbf{Exo$\to$Ego} \\
        \midrule
        DOMR & 48.0 & 51.5 & 48.8 & -- \\
        DOMR & 47.7 & 51.6 & -- & 53.3 \\
        \midrule
        DOMR$^\dagger$ & \textbf{49.0} & \textbf{53.4} & \textbf{49.7} & -- \\
        DOMR$^\dagger$ & \textbf{49.0} & \textbf{53.4} & -- & \textbf{55.2} \\
        \bottomrule
    \end{tabular}
\end{table}

\subsection{Qualitative Results}

To qualitatively demonstrate DOM's performance, we visualize representative matching results across diverse scenarios.

\begin{figure}[t]
    \centering
    \includegraphics[width=0.96\linewidth, trim=3.5cm 4.5cm 4cm 4.5cm, clip]{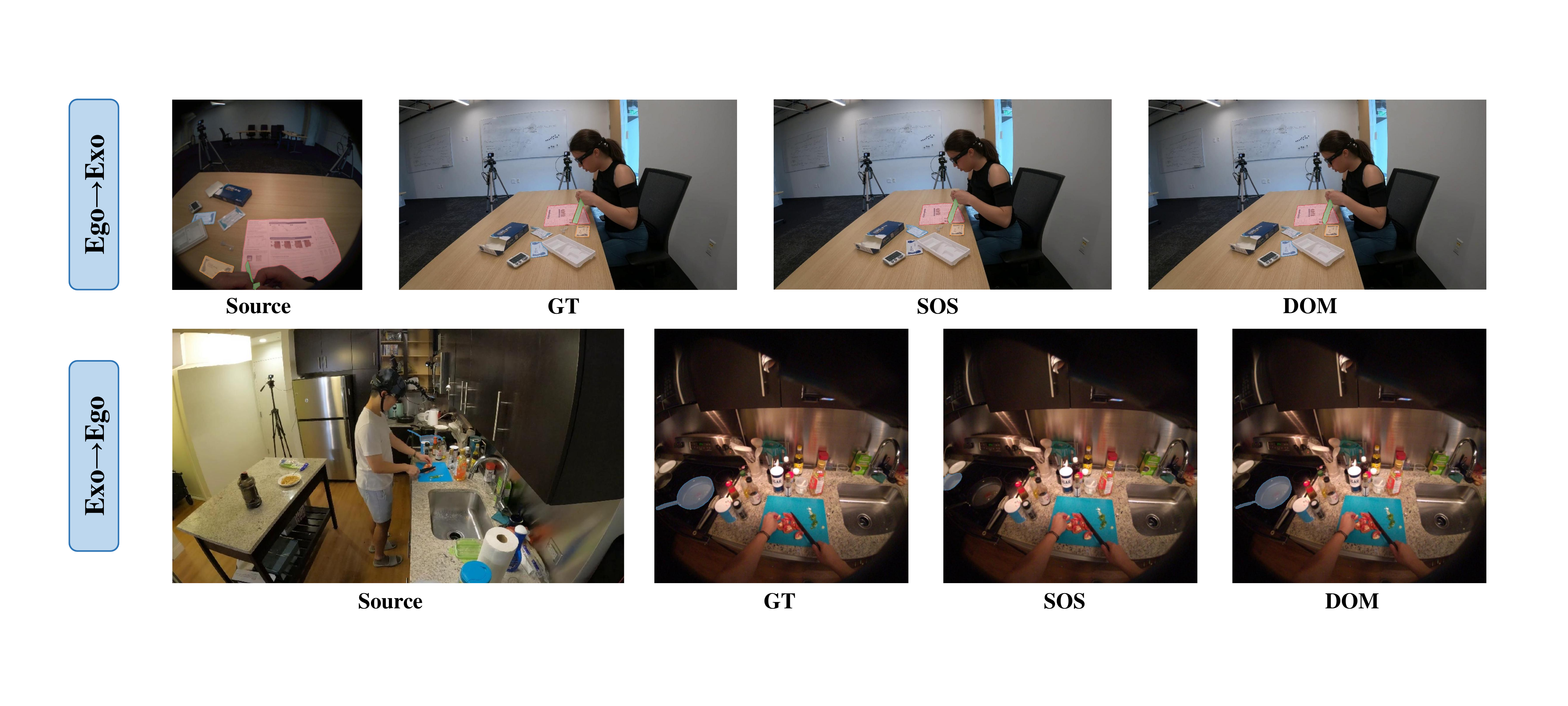}
    \caption{Comparative visualization of SOS and DOM performance on Ego$\to$Exo and Exo$\to$Ego tasks. The ``Source'' column shows the input image and its corresponding mask, while ``GT'' displays the target image with ground-truth. Distinct objects are highlighted with consistent colors in a row.}
    \label{fig:vis-dom}
\end{figure}

\noindent\textbf{Comparison between SOS and DOM}. 
Figure~\ref{fig:vis-dom} presents comparative results between SOS and DOM on both Ego$\to$Exo and Exo$\to$Ego tasks. In the Ego$\to$Exo scenario where multiple rectangular packages exist in the target image, SOS incorrectly matches the package masked in blue with a similar one when using a single query without referencing other objects. A similar situation occurs in the Exo$\to$Ego task: SOS fails to distinguish between two spatially adjacent cooking pots, resulting in incorrect matching. In contrast, DOM successfully handles these cases by leveraging reference relationships to correctly identify the target object among multiple confusing proposals. Comparison with ground truth results demonstrates DOM's effectiveness and accuracy in cross-view matching, providing reliable segmentation even in complex scenarios with multiple confusing objects.

\begin{figure}[t]
    \centering
    \hfill
    \begin{minipage}{0.46\linewidth}
        \includegraphics[width=\linewidth, trim=17cm 5.5cm 24cm 6cm, clip]{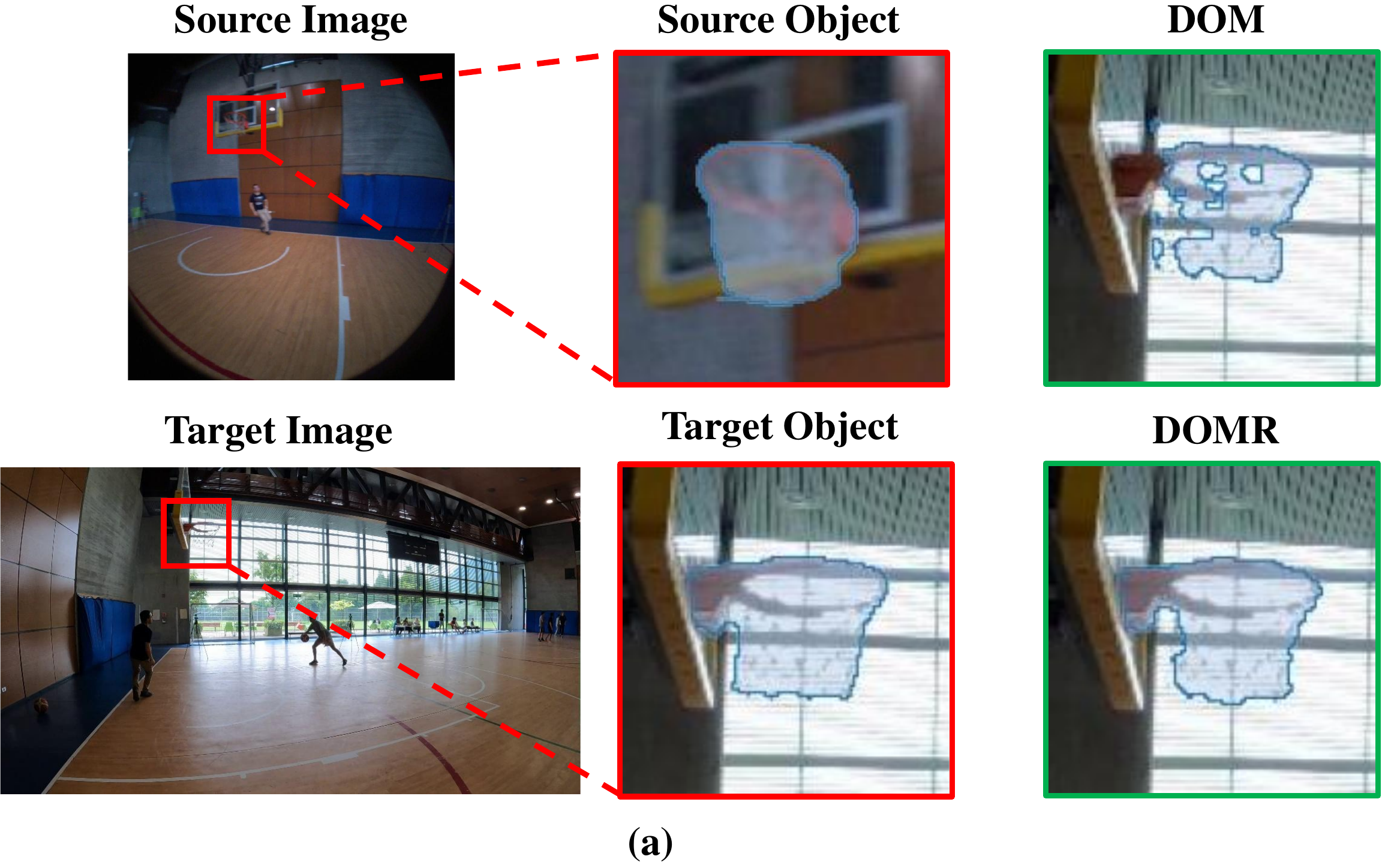}
    \end{minipage}
    \hfill
    \begin{minipage}{0.46\linewidth}
        \includegraphics[width=\linewidth, trim=17cm 5.5cm 24cm 6cm, clip]{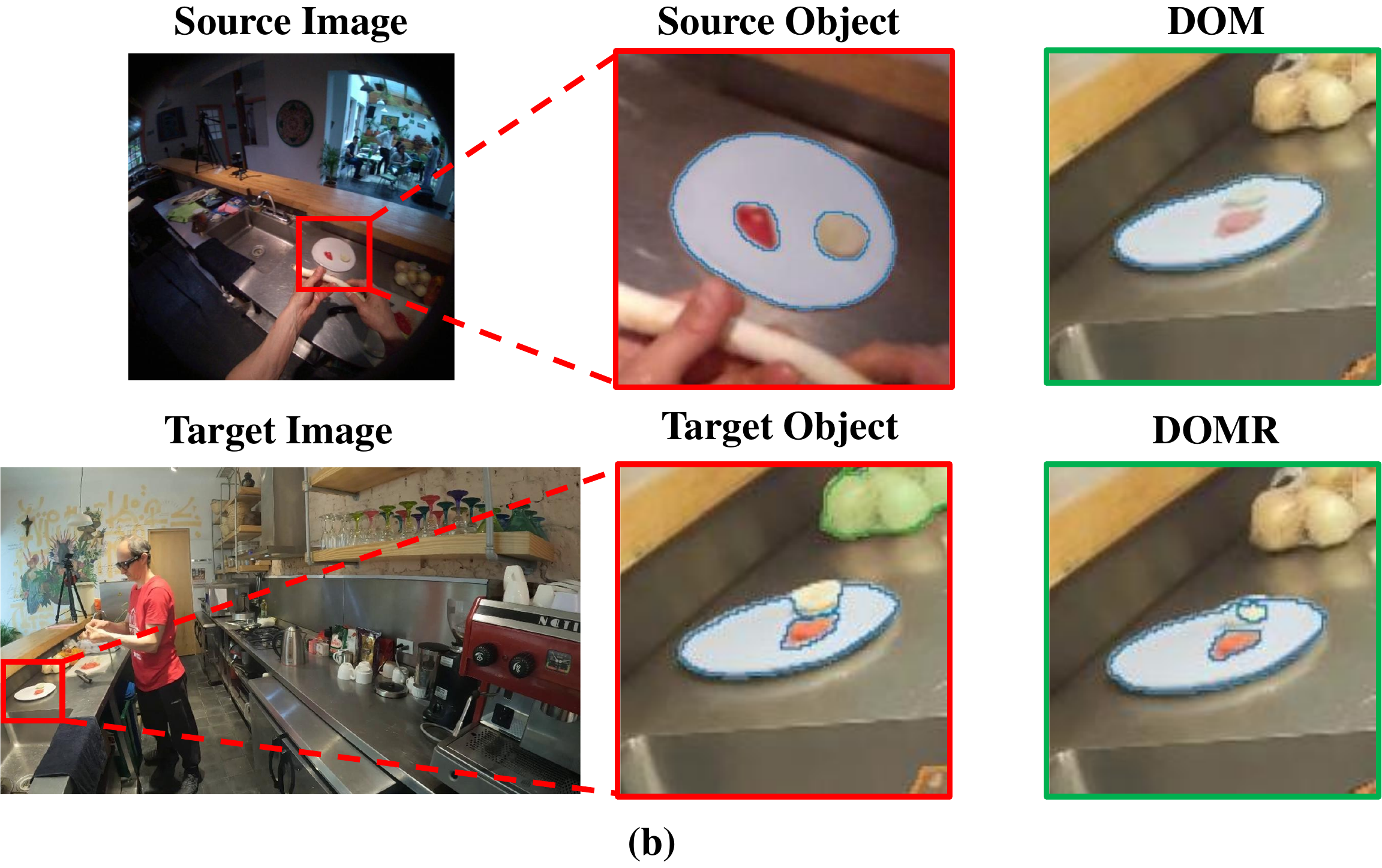}
    \end{minipage}
    \hfill
    \caption{Illustration of DOMR prediction before and after the MR module.}
    \label{fig:mhead}
\end{figure}

\noindent\textbf{Effectiveness of MR Module}. 
Bias accumulation during mask proposal generation can lead to incompleteness or inaccuracy in DOM, affecting final precision. Figure~\ref{fig:mhead} demonstrates the raw prediction masks before and after MR module processing. In the upper example, strong background lighting causes holes in the proposal mask, but MR completes the basketball net, forming a closed shape that aligns better with the ground truth. In the lower example, DOM successfully matches a complete plate with food on it, and MR further separates the plate from the food, achieving more accurate segmentation. These two cases effectively demonstrate how the MR module refines prediction results when dealing with suboptimal proposals, illustrating its effectiveness within the DOMR.

\section{Conclusion}

In this paper, we introduce the Dense Object Matching and Refinement (DOMR) framework for establishing dense object correspondences between egocentric and exocentric views. At its core is a Dense Object Matcher (DOM) module that jointly models multiple objects.
By jointly modeling visual features, spatial relationships and semantic contexts across viewpoints, DOM effectively captures complex inter-object dependencies that enable robust dense object matching. The framework is further enhanced by a dedicated mask refinement component that improves prediction accuracy through cross-view consistency optimization, forming the complete DOMR framework. Extensive evaluations on the Ego-Exo4D benchmark show that our approach achieves state-of-the-art performance, demonstrating significant improvements over existing methods. Although the ego-exo correspondence task is newly proposed and remains relatively underexplored, we believe our work provides a strong, comprehensive and reproducible baseline approach, and will inspire future research in this emerging direction.

\clearpage

\begin{acks}
This research is supported in part by National Key R\&D Program of China (2022ZD0115502), National Natural Science Foundation of China (NO.62461160308, U23B2010), “Pioneer” and “Leading Goose” R\&D Program of Zhejiang (No. 2024C01161).
\end{acks}

\bibliographystyle{ACM-Reference-Format}
\balance
\bibliography{refs}



\end{document}